\documentclass[10pt,twocolumn,letterpaper]{article}

\usepackage{amsmath,amsfonts,bm}

\def\eqref#1{equation~\ref{#1}}

\def\1{\bm{1}}

\def\rmA{{\mathbf{A}}}

\def\rmD{{\mathbf{D}}}

\def\rmF{{\mathbf{F}}}

\def\rmX{{\mathbf{X}}}

\def\vf{{\bm{f}}}

\def\vx{{\bm{x}}}

\DeclareMathAlphabet{\mathsfit}{\encodingdefault}{\sfdefault}{m}{sl}
\SetMathAlphabet{\mathsfit}{bold}{\encodingdefault}{\sfdefault}{bx}{n}

\usepackage{iccv}
\usepackage{times}
\usepackage{epsfig}
\usepackage{graphicx}
\usepackage{amsmath}
\usepackage{amssymb}
\usepackage{booktabs}

\usepackage[breaklinks=true,bookmarks=false]{hyperref}

\usepackage[capitalize,noabbrev]{cleveref}

\newcommand{\RN}[1]{%
    \textup{\lowercase\expandafter{\it \romannumeral#1}}%
}	

\iccvfinalcopy 

\def\iccvPaperID{2057} 

\ificcvfinal\fi

\begin{document}

\title{Learning Neural Eigenfunctions for Unsupervised Semantic Segmentation}

\author{Zhijie Deng\\
Shanghai Jiao Tong University\\
{\tt\small zhijied@sjtu.edu.cn}
\and
Yucen Luo\\
Max Planck Institute for Intelligent Systems\\
{\tt\small luoyucen9322@gmail.com}
}

\maketitle
\ificcvfinal\thispagestyle{empty}\fi

\begin{abstract}
Unsupervised semantic segmentation is a long-standing challenge in computer vision with great significance. Spectral clustering is a theoretically grounded solution to it where the spectral embeddings for pixels are computed to construct distinct clusters. Despite recent progress in enhancing spectral clustering with powerful pre-trained models, current approaches still suffer from inefficiencies in spectral decomposition and inflexibility in applying them to the test data. 
This work addresses these issues by casting spectral clustering as a parametric approach that employs neural network-based eigenfunctions to produce spectral embeddings. The outputs of the neural eigenfunctions are further restricted to discrete vectors that indicate clustering assignments directly. As a result, an end-to-end NN-based paradigm of spectral clustering emerges. In practice, the neural eigenfunctions are lightweight and take the features from pre-trained models as inputs, improving training efficiency and unleashing the potential of pre-trained models for dense prediction. We conduct extensive empirical studies to validate the effectiveness of our approach and observe significant performance gains over competitive baselines on Pascal Context, Cityscapes, and ADE20K benchmarks. 
\end{abstract}

\section{Introduction}
Semantic segmentation is essential in understanding the inherent structure and fine-grained information of images. However,
current approaches often hinge on a vast amount of manual annotations to train neural networks (NNs) effectively in an end-to-end manner~\cite{chen2018encoder,strudel2021segmenter}. This is problematic, as obtaining these annotations can be both time-consuming and costly, particularly in fields such as autonomous driving and medical image processing, where annotations are usually collected from domain experts. Thus, finding a way to perform semantic segmentation without manual annotation remains an important unresolved problem.

Unsupervised semantic segmentation has recently attracted a great deal of attention.
A number of methods attempt to tackle it by learning fine-grained image features using self-supervised objectives and then applying clustering or grouping techniques~\cite{zhang2020self,van2021unsupervised}. 
They tend to recognize single objects or single semantic categories and struggle with complex images. 
Other approaches have tried to use vision-language cross-modal models (e.g., CLIP~\cite{radford2021learning}) to achieve zero-shot semantic segmentation~\cite{zhou2022extract,shin2022reco}, but they heavily rely on carefully-tuned text prompts and self-training.
Compared to the recent approaches, the classic spectral clustering~\cite{shi2000normalized}, which has stood the test of time, remains an appealing option.
In particular, it enjoys solid foundations in  spectral graph theory---it finds the minimum cut of the connectivity graph over pixels. 

However, traditional spectral clustering exhibits limitations in three aspects: (\RN{1}) it operates on raw image pixels, thus is sensitive to color transformations and unable to recognize semantic similarities;
(\RN{2}) it is computationally inefficient due to the involved spectral decomposition; (\RN{3}) unlike NN-based methods, it is nontrivial and costly to extend to non-training samples because of its transductive manner.
Thus it cannot be performed in an end-to-end manner in the test phase.
Recent work reveals that pre-trained models such as ViTs~\cite{dosovitskiy2020image} can mitigate the first limitation and significantly improve the applicability and effectiveness of spectral clustering~\cite{melas2022deep}. 
Its core contribution is to build the connectivity graph over image patches based on an affinity matrix computed with the dense features from pre-trained models. 
Still, the limitations regarding efficiency and flexibility remain. 




The present paper aims to overcome the remaining limitations, rendering spectral clustering a simple yet effective baseline for unsupervised semantic segmentation.
To tackle the inefficiency issue, we propose to cast the involved spectral decomposition problem as an NN-based optimization one using the recently developed neural eigenfunction (NeuralEF) technique~\cite{deng2022neuralef}.  
Concretely, we first measure the similarities between image patches using both the features extracted from pre-trained models and raw pixels. 
Treating the similarity matrix (or its variants) as the output of a kernel function, we then optimize NNs to approximate its principal eigenfunctions. 
Consequently, our method constitutes an NN-based counterpart of spectral embedding. We eliminate the need for an additional grouping step, which is required in prior work~\cite{melas2022deep}, by constraining the NN output to one-hot vectors that indicate clustering assignments directly. To accomplish this, we use the Gumbel-Softmax estimator~\cite{jang2016categorical} for gradient-based optimization during training. 
These strategies transform spectral clustering from a non-parametric approach to a parametric one, enabling easy and reliable out-of-sample generalization and avoiding solving complex matrix eigenvalue problems during testing.

We perform extensive studies to evaluate the effectiveness of our approach for unsupervised semantic segmentation. 
We first experiment on the popularly used benchmarks Pascal Context~\cite{mottaghi2014role} and Cityscapes~\cite{cordts2016cityscapes} based on pre-trained ViTs, and report superior results compared to leading methods {MaskCLIP}~\cite{zhou2022extract} and {ReCo}~\cite{shin2022reco}. 
We further consider the sliding-window-based evaluation protocol~\cite{strudel2021segmenter} and experiment on the challenging ADE20K dataset~\cite{zhou2019semantic} to systematically study the behavior of our method. 
In addition, we conduct thorough ablation studies to gain insights into the specification of several core hyper-parameters.

\section{Related Work}
\noindent \textbf{Unsupervised segmentation.} Image segmentation has numerous practical applications in various industries and scientific fields.
In order to alleviate the burden of collecting annotations, prior works have comprehensively studied the problem of learning image segmenters in semi- and weakly-supervised settings~\cite{khoreva2017simple,huang2018weakly,chan2021comprehensive,ke2021universal}. 
Recently, there have been increasing efforts to tackle unsupervised segmentation based on the progress in related fields like deep generative models (DGMs) and self-supervised learning (SSL). 
On the one hand, DGM-based segmentation approaches train specialized image generators to separate foreground from background~\cite{chen2019unsupervised,arandjelovic2019object,abdal2021labels4free} or extract saliency masks directly from pre-trained generators~\cite{voynov2021object}. 
Yet, it is technically non-trivial to extend them to cope with \emph{semantic} segmentation. 
On the other hand, SSL-based methods define objectives to perform clustering~\cite{hwang2019segsort,cho2021picie}, mutual information maximization~\cite{ji2019invariant,ouali2020autoregressive}, contrastive learning~\cite{van2021unsupervised} or feature correspondence distillation~\cite{hamilton2022unsupervised} to learn image features suitable for grouping. 
However, most of these methods only recognize single objects or single semantic categories and struggle with complex images. 
With the increasing accessibility of pre-trained image-text cross-modal models, considerable efforts have been devoted to performing zero-shot semantic segmentation with them~\cite{zhou2022extract,shin2022reco}.
Nevertheless, the entanglement with cross-modal models places high demands on the quality of the text prompts, which correspond to the semantic categories of concern, and hinders the methods from choosing backbone models freely.

\noindent \textbf{Spectral clustering.} 
As a classic solution to image segmentation, spectral clustering~\cite{shi2000normalized,ng2001spectral} frames the original problem as a graph partitioning one defined on the connectivity graph over image pixels. 
Typically, spectral clustering exploits the eigenvectors of graph Laplacians to construct minimum-energy graph partitions~\cite{donath1973lower,fiedler1973algebraic}. Spectral clustering is closely related to Kernel PCA~\cite{scholkopf2005kernel} as they are both learning eigenfunctions~\cite{bengio2003learning}.
Recently, spectral clustering has been combined with pre-trained models to enjoy rich semantic information~\cite{melas2022deep}. 
Yet, it remains unsolved that spectral decomposition is expensive for big data, and the non-parametric nature hinders out-of-sample generalization. 
This paper aims to address these issues.

\noindent \textbf{The deep learning variant of spectral methods. }
Refurbishing spectral methods with deep learning techniques is beneficial to improve the scalability and flexibility of the former. 
The spectral inference networks (SpIN)~\cite{pfau2018spectral} is a seminal work in this direction. 
Yet, the learning objective of SpIN is ill-defined, which leads to only the subspace spanned by the principal eigenfunctions instead of the eigenfunctions themselves. 
SpIN hence introduces convoluted and expensive strategies to solve this problem. 
The recent NeuralEF technique~\cite{deng2022neuralef} alternatively defines a new series of objective functions to explicitly break the symmetry among eigenfunctions. 
NeuralEF is further enhanced by weight sharing and extended to handle indefinite kernels in \cite{deng2022neural}, constituting a more amenable choice for learning spectral embeddings given pre-defined kernels. 
We also note that there are several attempts to develop deep spectral clustering methods based on supervisions~\cite{law2017deep} or a dual autoencoder network~\cite{yang2019deep}, but they cannot be trivially applied to the task of unsupervised semantic segmentation due to the absence of annotations or other inefficiency issues.

\section{Methodology}
\begin{figure*}[t]
\begin{center}
\includegraphics[width=\linewidth]{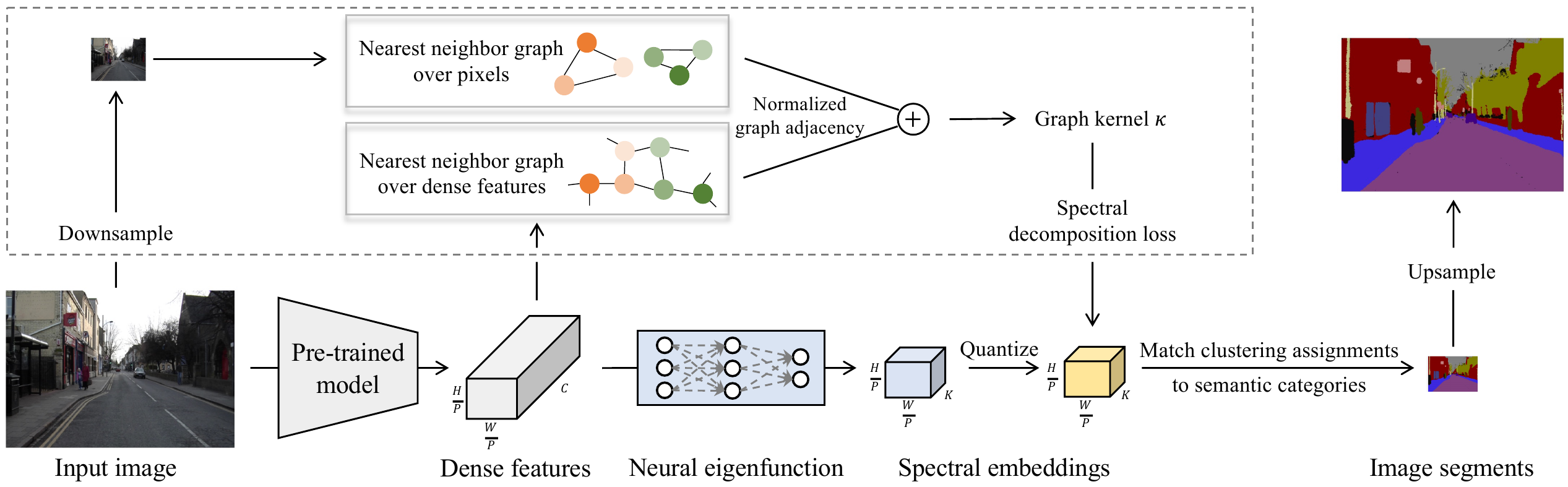}
\end{center}
   \caption{Method overview. We establish an end-to-end NN-based pipeline for spectral clustering to perform unsupervised semantic segmentation. The box highlights modules that exist only during training. The nearest neighbor graph over pixels only has edges between pixels from the same images, while the other one is built over dense features from various images. The pre-trained model is \emph{fixed}. 
   }
\label{fig:1}
\vspace{-2ex}
\end{figure*}

We begin with a brief review of the relevant background and then build up our approach step by step.
We provide an overview of the proposed method in \cref{fig:1}. 

\subsection{Background}

Semantic segmentation essentially targets determining the semantic consistency among image pixels. 
In the absence of annotations, the problem boils down to an unsupervised clustering one. 
Spectral clustering~\cite{shi2000normalized,ng2001spectral} is a long-standing solution to it with solid theoretical foundations. 
It proceeds by partitioning a connectivity graph over image pixels.  
The resulting algorithm typically involves eigendecomposing the graph Laplacian matrix and stacking its eigenvectors, which are then used in a Euclidean clustering algorithm to obtain a fixed number of partitions.

Traditional spectral clustering operates on raw image pixels, making it sensitive to color transformations and unable to identify semantic similarities.
To address this, a natural idea is to include the inductive bias of NNs. 
In this spirit, the deep spectral method (DSM)~\cite{melas2022deep} leverages powerful pre-trained models to translate the learning from raw pixels to patch-wise features that embed rich local semantics and have been proven to be widely applicable. 
Specifically, let $\vx_i \in \mathbb{R}^{H\times W \times 3}$ denote an image and $f: \mathbb{R}^{ H\times W \times 3} \to \mathbb{R}^{ H/P \times W/P \times C}$ denote the pre-trained model with $P$ as a down-sampling factor. 
DSM constructs a semantic affinity matrix using both the patch-wise features $\vf_{i, j,k}=f(\vx_i)_{j,k} \in \mathbb{R}^{C}$ and image pixels, and performs spectral clustering on the corresponding connectivity graph. 
Subsequently, DSM performs bi-linear interpolation to convert patch-wise segments to pixel-wise ones.
This simple workflow has surpassed various competitors regarding unsupervised semantic segmentation performance. 

Denote by $N$ the size of the dataset of concern.
Semantic segmentation requires classifying pixels in different images yet with the same semantics into the same category. 
In other words, it needs to ensure semantic consistency across the dataset.
Therefore, DSM should, in principle, work on an affinity matrix of size $\mathbb{R}^{NHW/P^2 \times NHW/P^2}$, 
but this is computationally infeasible as the involved spectral decomposition has a cubic complexity w.r.t. $NHW/P^2$. 
DSM alternatively conducts spectral clustering separately for each image and then performs cross-image synchronization, but this is obviously inflexible compared to NN-based pipelines and arguably suboptimal. Additionally, existing spectral clustering methods, including DSM, are non-parametric, relying on costly spectral decomposition when faced with new test data. 

In order to address the limitations discussed above, we leverage a parametric approach to spectral decomposition based on the recent NeuralEF technique~\cite{deng2022neuralef} and learn discrete neural eigenfunctions directly. 
These innovations help to scale our method to large datasets, enable the straightforward and cheap out-of-sample extension for test data, and establish an NN-based spectral clustering workflow.

\subsection{From Eigenvectors to Eigenfunctions}

Spectral clustering involves the spectral decomposition of a matrix and hence is seemingly incompatible with the parametric NN models. 
To bridge this gap, we move our viewpoint from the eigenvectors to eigenfunctions.

Specifically, abstracting the graph Laplacian matrix as the evaluation of some kernel function $\kappa$ on the dense features, the eigenfunctions of $\kappa$ form a function-space generalization of the aforementioned eigenvectors. 
Formally, the eigenfunction $\psi$ of a kernel $\kappa(\vx, \vx')$ satisfies that
\begin{equation}
    \int \kappa(\vx, \vx')\psi(\vx')p(\vx') d\vx' = \mu \psi(\vx),
\end{equation}
where $\mu$ is the corresponding eigenvalue and $p$ a probability measure. 
By definition, the eigenfunction should be normalized, and different eigenfunctions are orthogonal. 

As shown, the eigenfunction takes input from the original space and maps it to the eigenspace specified by the kernel, where the local neighborhoods on data manifolds are preserved. 
Naturally, we can incorporate NNs as function approximators to the eigenfunctions. 
In this way, we bypass the need for expensive spectral decomposition of a matrix and can easily perform out-of-sample extension thanks to the generalization ability of NNs. 
Fortunately, the recently proposed NeuralEF~\cite{deng2022neuralef} offers tools to realize this. 
We also note that the effectiveness of the spectral embedding yielded by neural eigenfunctions has been empirically validated in self-supervised learning~\cite{deng2022neural}.

\subsection{Setting up the Graph Kernel}
To set up the kernel for spectral clustering,
as suggested by DSM~\cite{melas2022deep}, we construct two graphs with patch-wise features from pre-trained models and down-sampled image pixels, respectively.
We then define the kernel with the weighted sum of the corresponding normalized graph adjacency matrices. 
By doing so, the high-level semantics and low-level details are conjoined. 

Denote by $\rmF = \{\vf_i\}_{i=1}^N \in \mathbb{R}^{NHW/P^2 \times C}$ the collection of patch-wise features over the dataset. 
Considering that the feature space shaped by large-scale pre-training is highly structured where simple distance measures suffice to represent similarities, we leverage cosine similarity to specify a nearest neighbor graph over image patches. 
The corresponding affinity matrix is detailed below:
\begin{equation}
  \rmA_{u,v} =
    \begin{cases}
      {\rmF_u^{}} {\rmF_v^\top}/({\Vert \rmF_u \Vert}{\Vert \rmF_v \Vert}), & v \in {\text{$k$-NN}}(u, \rmF, \text{cosine})\\
      0 & \text{otherwise}
    \end{cases}       
\end{equation}
where ${\text{$k$-NN}}(u, \rmF, \text{cosine})$ denotes the set of the $k$ nearest neighbors of $\rmF_u$ over $\rmF$ under cosine similarity. 
The above graph deploys edges between patches that are semantically close.
Thereby, the corresponding graph partitioning can result in meaningful segmentation at a coarse resolution. 
Although other traditional graphs can also be used, the nearest neighbor graph enjoys sparsity, which reduces storage and computation costs, and has been studied extensively in manifold learning and spectral clustering.

To supplement the high-level features with low-level details,
we bilinearly down-sample the original image $\vx_i$ to $\tilde{\vx}_i \in \mathbb{R}^{H/P \times W/P \times 3}$ to keep resolution consistency, and fuse the spatial information $(j, k)$ and color information $\tilde{\vx}_{i, j, k}$ as a single vector. 
We stack the collection over the dataset as $\tilde{\rmX} \in \mathbb{R}^{NHW/P^2 \times 5}$. 
A nearest neighbor graph over down-sampled pixels is then defined based on $L^2$ distance following DSM~\cite{melas2022deep}. 
The affinity matrix is:
\begin{equation}
    \tilde{\rmA}_{u,v} =
    \begin{cases}
      1, & v \in {\text{$\tilde{k}$-NN}}(u, \tilde{\rmX}, L^2)\\
      0 & \text{otherwise}
    \end{cases} 
\end{equation}
where ${\text{$\tilde{k}$-NN}}(u, \tilde{\rmX}, L^2)$ denotes the set of $\tilde{k}$ nearest neighbors of $\tilde{\rmX}_u$ over $\tilde{\rmX}$ under $L^2$ distance (dissimilarity). 
We place a further constraint that the nearest neighbors should belong to the same image as the query to avoid establishing meaningless connections. 
It is expected that such a graph helps to detect sharp object boundaries.

We symmetrize $\rmA$ and $\tilde{\rmA}$ so that they can serve as graph adjacency matrices. 
Considering that the normalized graph cuts are more practically useful~\cite{shi2000normalized}, we would better build a kernel with the normalized graph Laplacian matrices of $\rmA$ and $\tilde{\rmA}$, and learn the eigenfunctions associated with the smallest $K$ eigenvalues. 
However, the NeuralEF approach instead deploys neural approximations to the \emph{principal} eigenfunctions, which correspond to the largest eigenvalues, of a kernel~\cite{deng2022neuralef}. 
To this end, we move our focus from the normalized Laplacian matrix to the normalized adjacency matrix whose $K$ principal eigenfunctions exactly correspond to the eigenfunctions associated with the $K$ smallest eigenvalues of the former. 
Specifically, let $\rmD:=\mathrm{diag}(\rmA \mathbf{1}), \tilde{\rmD}:=\mathrm{diag}(\tilde{\rmA} \mathbf{1})$ denote the degree matrices. 
We then define the kernel function $\kappa$ using the weighted sum of the normalized adjacency matrices, detailed below:
\begin{equation}
    \kappa : \kappa(\rmF, \rmF) =  {\rmD}^{-1/2}\rmA {\rmD}^{-1/2} + \alpha \tilde{\rmD}^{-1/2}\tilde{\rmA} \tilde{\rmD}^{-1/2},
\end{equation}
where $\alpha$ is a trade-off parameter. 
Here we define the kernel in the space of features extracted from pre-trained models while other choices are also viable: it only affects the input of neural eigenfunctions. 
In the current setting, the features from pre-trained models are directly fed into neural eigenfunctions, which is sensible thanks to the great potential of pre-trained models and can prevent unnecessary training costs.



\subsection{Learning Neural Eigenfunctions}

Let $\bm{\mathrm{f}} \in \mathbb{R}^C$ denote a row vector of $\rmF$ and $p(\bm{\mathrm{f}})$ a uniform distribution over $\{\rmF_i\}_{i=1}^{NHW/P^2}$. 
The NeuralEF technique~\cite{deng2022neuralef} approximates the $K$ principal eigenfunctions of the kernel $\kappa$ w.r.t. $p(\bm{\mathrm{f}})$ with a $K$-output neural function $\psi: \mathbb{R}^{C} \to \mathbb{R}^K$ by solving the following problem:
\begin{equation}
\small
\label{eq:1}
\max_\psi \sum_{j=1}^K \mathbf{R}_{j,j} - \beta \sum_{j=1}^K \sum_{i=1}^{j-1} \widehat{ \mathbf{R}}_{i,j}^2, \; s.t.\; \mathbb{E}_{p(\bm{\mathrm{f}})}[\psi(\bm{\mathrm{f}}) \circ \psi(\bm{\mathrm{f}})] = \mathbf{1},
\end{equation}
where $\beta$ is a positive coefficient and $\circ$ is Hadamard product, and
\begin{equation}
\small
\begin{aligned}
\mathbf{R} &:= \mathbb{E}_{p(\bm{\mathrm{f}})}\mathbb{E}_{p(\bm{\mathrm{f}}')} [{\kappa}(\bm{\mathrm{f}}, \bm{\mathrm{f}}')\psi(\bm{\mathrm{f}}) \psi(\bm{\mathrm{f}}')^\top] \\
\widehat{ \mathbf{R}} &:= \mathbb{E}_{p(\bm{\mathrm{f}})}\mathbb{E}_{p(\bm{\mathrm{f}}')} [{\kappa}(\bm{\mathrm{f}}, \bm{\mathrm{f}}') \hat{\psi}(\bm{\mathrm{f}}) \psi(\bm{\mathrm{f}}')^\top].
\end{aligned}
\end{equation}
Here $\hat{\psi}$ is the non-optimizable variant of $\psi$. 

Adapting such results to our case and applying Monte Carlo estimation to the expectation yields
\begin{equation}
\small
\label{eq:R}
\mathbf{R} \approx \frac{1}{B^2} \mathbf{\Psi} \cdot  \kappa(\rmF^B, \rmF^B) \cdot  \mathbf{\Psi}^\top \; \text{and} \; \widehat{\mathbf{R}} \approx \frac{1}{B^2} \widehat{\mathbf{\Psi}} \cdot  \kappa(\rmF^B, \rmF^B) \cdot  \mathbf{\Psi}^\top,
\end{equation}
where $\rmF^B := [\bm{\mathrm{f}}_1,\dots, \bm{\mathrm{f}}_B]^\top \in \mathbb{R}^{B \times C}$ is a mini-batch of features, $\mathbf{\Psi}:=[\psi(\bm{\mathrm{f}}_1),\dots, \psi(\bm{\mathrm{f}}_B)] \in \mathbb{R}^{K \times B}$ is the collection of the corresponding NN outputs, and $\widehat{\mathbf{\Psi}} := \mathrm{stop\_gradient}(\mathbf{\Psi})$. 
The $\mathrm{stop\_gradient}$ operation can be easily found in auto-diff libraries and plays a vital role in establishing the asymmetry between the $K$ principal eigenfunctions. 
We put an $L^2$-batch normalization layer~\cite{deng2022neuralef} at the end of model $\psi$ to enforce the constraint in \cref{eq:1}. 

$\psi$ can be implemented as a neural network composed of convolutions or transformer blocks, which takes dense image features $\vf=f(\vx) \in \mathbb{R}^{H/P \times W/P \times C}$ as input and outputs $\mathbf{\psi}(\vf) \in \mathbb{R}^{H/P \times W/P \times K}$. 
After training, it can be used to predict new test data without relying on expensive test-time spectral decomposition. 

\subsection{Quantizing Neural Eigenfunctions}
After obtaining the spectral embedding $\psi(f(x))$ for image patches, we should, by convention, invoke a Euclidean clustering algorithm such as K-means~\cite{lloyd1982least} to get clustering assignments. 
In practice, datasets can often be large, so it would be better to leverage online clustering mechanisms to avoid incurring unaffordable storage costs. 
However, this approach may still be time-consuming when aiming for good convergence. Additionally, this pipeline is not as flexible as NN-based segmenters.

To bridge the gap, we impose a constraint on the output of $\psi$ to be $K$-dim one-hot vectors which directly indicate clustering assignments.
We resort to the Gumbel-Softmax estimator~\cite{jang2016categorical} for gradient-based optimization. 
This follows the notion that the outputs of eigenfunctions are soft clustering assignments and thus we further perform quantization of them. 
This is also in a similar spirit to the spectral hashing approach~\cite{weiss2008spectral} where the output of $\psi$ is assumed to be vectors over $\{-1, 1\}^K$. 
We clarify that the Gumbel-Softmax estimator precedes the aforementioned $L^2$ batch normalization layer. 
In the test phase, we remove this estimator and the $L^2$ batch normalization layer, using the NN outputs directly as softmax logits for clustering. This results in a pure NN-based workflow for spectral clustering.

\subsection{From Clusters to Image Segments}

During inference, we first up-sample the softmax logits bilinearly to match the original resolution of input images. We then apply the argmax operation to obtain discrete clustering assignments. 
A natural question that arises is how to match these clustering assignments to pre-defined semantics to obtain the final image segments. Two well-studied solutions are
Hungarian matching~\cite{kuhn1955hungarian} and majority voting. As per DSM~\cite{melas2022deep}, when evaluated on standard image semantic segmentation benchmarks, we first collect the clustering assignments for all validation images then match them to ground-truth labels via these approaches to conduct a quantitative evaluation of segmentation performance.

\section{Experiments}
To evaluate the efficacy of the proposed method for unsupervised semantic segmentation, we conduct comprehensive experiments on various standard benchmarks. 

\subsection{Experimental Setups}

\noindent \textbf{Datasets.} We primarily experiment on Pascal Context and Cityscapes, which consist of $60$ and $27$ classes, respectively. We employ the same data pre-processing and augmentation strategies as the work~\cite{strudel2021segmenter} on the training dataset. 

\noindent \textbf{Pre-trained models.} We use pre-trained ViTs~\cite{dosovitskiy2020image} provided by the timm library~\cite{rw2019timm} and consider the ``Small'', ``Base'', and ``Large'' variants. 
The weights have been pre-trained on ImageNet-21k~\cite{ridnik2021imagenet21k} and fine-tuned on ImageNet~\cite{deng2009imagenet} in a supervised manner. 
In particular, we consider pre-trained models at a high resolution, like $384\times 384$, to obtain fine-grained segments. 
We fix the patch size to $16 \times 16$ (i.e., $P=16$) to better trade-off efficiency and accuracy. 
We use the output of the last attention block of ViTs as $\vf_i$ without elaborate selection. 
We also feed the intermediate features of the pre-trained ViTs, which can be freely accessed, to the neural eigenfunctions to enrich the input information.

\noindent \textbf{Modeling and training.} We set $k=256$ for the nearest neighbor graph defined on pre-trained models' features. To reduce the cost of searching for the nearest neighbors, we confine the search to the current mini-batch, which is shown to be empirically effective.
We specify the other graph following DSM~\cite{melas2022deep}. 
The trade-off coefficient $\alpha$ equals $0.3$ based on an ablation study reported in \cref{sec:abl}. 
The training objective is detailed in \cref{eq:1} and unfolded in \cref{eq:R}. 
As $K$ implies the number of semantic classes uncovered automatically, we make it larger than the number of ground-truth semantic classes and, in practice, set it to $256$. 
We set the trade-off coefficient $\beta$ to $0.08$ and linearly scale it for other values of $K$ (see the study in \cref{sec:abl}). 
We use $2$ transformer blocks with linear self-attention~\cite{katharopoulos-et-al-2020} and a linear head to specify the neural eigenfunctions $\psi$ for efficiency. 
We restrict the weight matrix of the linear head to have orthogonal columns and find it beneficial to the final performance empirically. 
We anneal the temperature for Gumbel-Softmax from $1$ to $0.3$ following a cosine schedule during training. 
The training relies on an Adam optimizer~\cite{kingma2014adam} and a learning rate of $10^{-3}$ (with cosine decay). 
No weight decay is employed. 
The training lasts for $40$ epochs with batch size $16$ (or $8$ if an out-of-memory error occurs), which takes half a day on one RTX 3090 GPU. 

\noindent \textbf{Evaluation.} We test on the validation set of the datasets and report both pixel accuracy (Acc.) and mean intersection-over-union (mIoU). 
We follow the common practice in unsupervised semantic segmentation where the validation images are resized and cropped to have $320\times 320$ pixels (see \cite{hamilton2022unsupervised,shin2022reco}).
In particular, the background class in the Pascal Context is excluded from evaluation.
We apply a softmax operation to the outputs of $\psi$ to obtain clustering assignments. 
We then use majority voting to match the resulting clusters to semantic segments. 
We apply CRF~\cite{krahenbuhl2011efficient} for post-processing (although it is empirically shown that its contribution to performance gain is marginal).

\begin{table}[t]
  \centering
  \begin{tabular}{lcc@{}}
  \toprule
Method  & \multicolumn{1}{c}{\quad\quad\textbf{Acc.} (\%)\quad\quad}
& \multicolumn{1}{c}{\textbf{mIoU} (\%)}\\ 
\midrule
\emph{MaskCLIP~\cite{zhou2022extract}} & - & 25.5\\
\emph{MaskCLIP+~\cite{zhou2022extract}} & - & 31.1\\
\emph{ReCo~\cite{shin2022reco}} & 51.6 & 27.2\\
\emph{K-means} \\
\quad \quad {\small ViT-S} & 61.9 & 28.9 \\
\quad \quad {\small ViT-B} & 58.7 & 30.9 \\
\quad \quad {\small ViT-L} & 45.3 & 19.3 \\
\emph{Ours} \\
\quad \quad  {\small ViT-S} & {70.4} & {38.8} \\
\quad \quad  {\small ViT-B} & 69.7 & 37.5 \\
\quad \quad  {\small ViT-L} & 63.2 & 33.2 \\
\emph{Ours*} \\
\quad \quad  {\small ViT-S} & \textbf{74.6} & \textbf{39.6} \\
\quad \quad  {\small ViT-B} & 73.2 & 37.6 \\
\quad \quad  {\small ViT-L} & 71.9 & 35.0 \\
  \bottomrule
   \end{tabular}
\vspace{1ex}
\caption{Comparisons on unsupervised semantic segmentation performance on Pascal Context~\cite{mottaghi2014role}. The results of MaskCLIP and ReCo are from the original papers.}
\vspace{-0.2cm}
  \label{table:1}
\end{table}

\subsection{Comparison with Leading Methods}
\label{sec:sota}
We first compare the proposed method to leading unsupervised semantic segmentation methods. 
We consider two representative competitors from the literature: {MaskCLIP}~\cite{zhou2022extract} and {ReCo}~\cite{shin2022reco}, which empirical outperform a variaty of baselines such as IIC~\cite{ji2019invariant}, PiCIE~\cite{cho2021picie}, and STEGO~\cite{hamilton2022unsupervised}. 
We also try to include the self-training variants of MaskCLIP and ReCo in comparison. 
Besides, we introduce two other baselines: (\RN{1}) fit a K-means with the features of pre-trained models for training data (using the same number of clustering centers as our method) and use it to predict clustering assignments for validation data; 
(\RN{2}) likewise, fit a K-means with the features preceding the linear head in $\psi$.
The two ways are referred to as ``K-means'' and ``Ours*'' in our studies. 
We have not compared with DSM~\cite{melas2022deep} because the involved divide-and-conquer procedure for synchronizing the clustering results across different images is non-trivial to implement. In theory, DSM can lead to a similar segmentation performance to our method (yet with more resource consumption). 

We report the results in \cref{table:1} and \cref{table:2}. As shown, our methods outperform MaskCLIP, ReCo, and K-means with significant margins, which reflects the efficacy of the learned neural eigenfunctions for unsupervised semantic segmentation.
``Ours*'' even outperforms ``Ours'', which is probably attributed to that the features preceding the linear head in $\psi$ have a much higher dimension than the final outputs and thus are more informative. 
While MaskCLIP+ and ReCo+ employ an extra time-consuming self-training step, they are still inferior to our methods.
It is reasonable to speculate that combining our methods with self-training can further improve performance. 
Note also that K-means can serve as a strong baseline for unsupervised semantic segmentation. This finding appears to be contrary to the results reported in DSM~\cite{melas2022deep} (Table 4). 
We deduce the reason that DSM sets the number of clusters to the number of true classes rather than a larger quantity and performs K-means directly on the validation set instead of the training set.
Besides, as the size of the pre-trained model increases, both K-means and our methods yield worse unsupervised segmentation performance. 
We attribute this to the fact that larger pre-trained models tend to produce more abstract outputs that contain more overall semantics than specialized details.
In contrast, smaller pre-trained models may fall short in expressiveness. Therefore, we suggest using a pre-trained model with appropriate capacity when applying the proposed method to new applications.

\begin{figure}[t]
\begin{center}
\includegraphics[width=\linewidth]{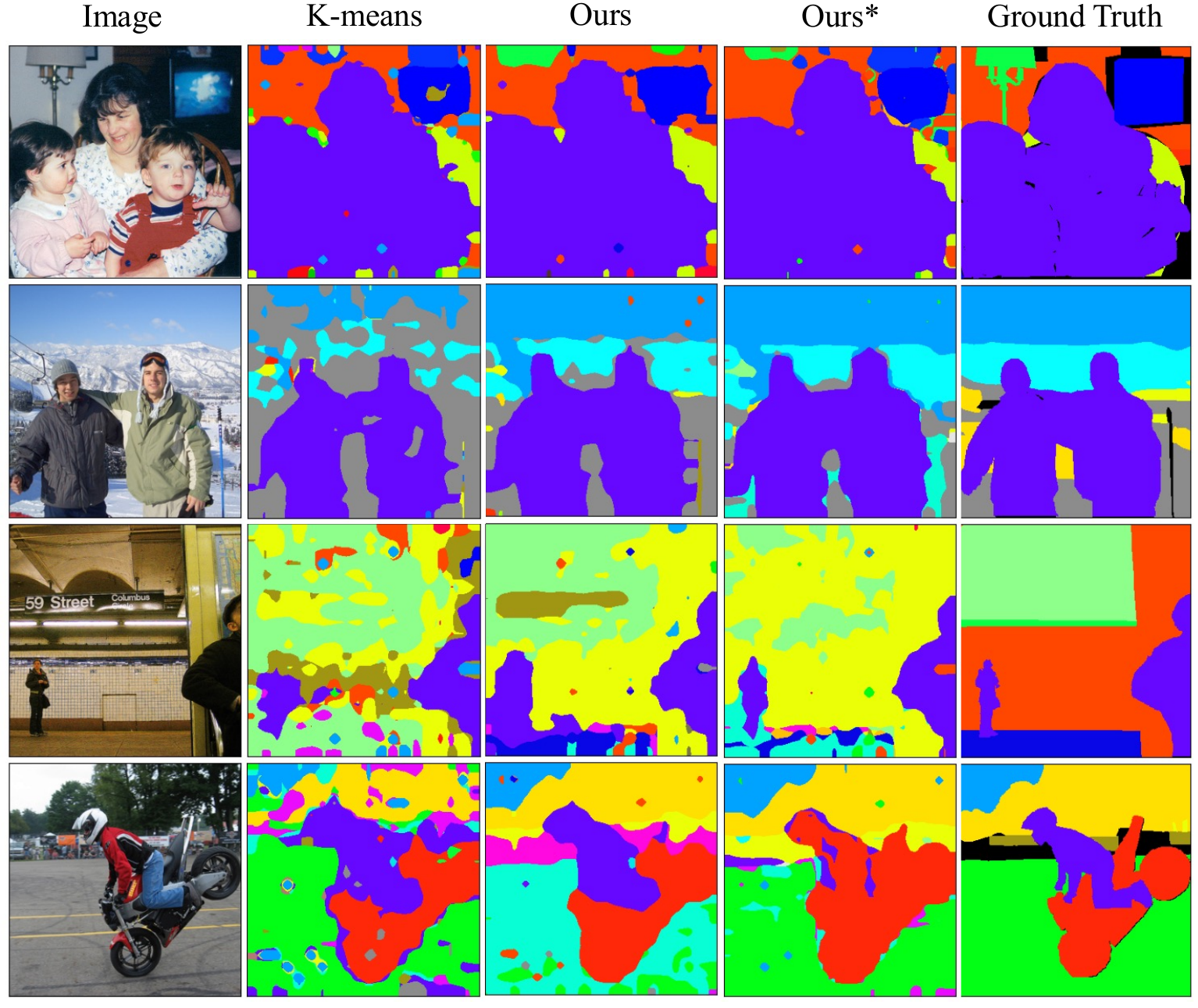}
\end{center}
\vspace{-.5ex}
   \caption{Visualization of the unsupervised semantic segmentation results on Pascal Context~\cite{mottaghi2014role}.}
  \vspace{-.5ex}
\label{fig:3}
\end{figure}

\noindent \textbf{Qualitative results.} \cref{fig:3} shows some qualitative results of the proposed methods on Pascal Context.
We also include the K-means baseline in comparison. 
Most notably, unlike MaskContrast~\cite{van2021unsupervised}, our method can detect multiple semantic categories in the same image, and the generated object boundaries are sharp. 
Furthermore, our spectral clustering-based approach can greatly reduce the chaotic fragments produced by K-means.
We defer the visualization for the learned neural eigenfunctions and the segmentation results on Cityscapes to Appendix.

\begin{table}[t]
  \centering
  \begin{tabular}{lcc@{}}
  \toprule
Method  & \multicolumn{1}{c}{\quad\quad\textbf{Acc.} (\%)\quad\quad}
& \multicolumn{1}{c}{\textbf{mIoU} (\%)}\\ 
\midrule
\emph{MaskCLIP~\cite{zhou2022extract}} & 35.9 & 10.0\\
\emph{ReCo~\cite{shin2022reco}} & 74.6 & 19.3\\
\emph{ReCo+~\cite{shin2022reco}} & 83.7 & 24.2\\
\emph{K-means} \\
\quad \quad {\small ViT-S} & 77.0 & 22.4 \\
\quad \quad {\small ViT-B} & 74.8 & 23.2 \\
\quad \quad {\small ViT-L} & 66.3 & 20.9 \\
\emph{Ours} \\
\quad \quad  {\small ViT-S} & {83.4} & {28.2} \\
\quad \quad  {\small ViT-B} & 81.4 & 26.8 \\
\quad \quad  {\small ViT-L} & 80.3 & 26.3 \\
\emph{Ours*} \\
\quad \quad  {\small ViT-S} & \textbf{84.6} & {30.0} \\
\quad \quad  {\small ViT-B} & 84.2 & \textbf{30.7} \\
\quad \quad  {\small ViT-L} & 84.2 & 30.0 \\
  \bottomrule
   \end{tabular}
\vspace{1ex}
\caption{Comparisons on unsupervised semantic segmentation performance on Cityscapes~\cite{cordts2016cityscapes}. The results of MaskCLIP and ReCo are from the original papers. }
  \label{table:2}

\vspace{-0.2cm}
\end{table}

\subsection{Evaluation in More Scenarios}
In this subsection, we evaluate the proposed method in more scenarios to systematically study its behavior. 

First, we consider the evaluation protocol popularly used in the \emph{supervised} setting.
Specifically, following the setup~\cite{strudel2021segmenter}, we leverage a sliding window with the same resolution as training images to cope with the varying sizes of evaluation images. 
The background class in the Pascal Context is included for evaluation, and the results on Cityscapes cover only $19$ categories. 
Due to implementation challenges, we have not included existing methods in comparison, but we introduce a supervised semantic segmentation baseline, DeepLabv3+~\cite{chen2018encoder}, which can serve as an upper bound on performance. 

As shown in \cref{table:3}, in this new evaluation setting, our methods can clearly outperform K-means, and ``Ours*'' is slightly better than ``Ours'' in general. 
These results are consistent with those reported in \cref{sec:sota} and help to verify the extensibility of the proposed method. 
We also notice that the performance gap between the proposed \emph{unsupervised} segmentation method and the \emph{supervised} baseline is not big, especially on the Pascal Context dataset, which largely motivates further investigation on the direction of improving spectral clustering.

\begin{table}[t]
  \centering
  \begin{tabular}{lcc@{}}
  \toprule
Method  & \multicolumn{1}{c}{\quad\quad\textbf{Acc.} (\%)\quad\quad}
& \multicolumn{1}{c}{\textbf{mIoU} (\%)}\\ 
\midrule
\textbf{Pascal Context} \\
\emph{Supervised} & - & \underline{48.5}\\
\emph{K-means} & 56.3 & 31.9 \\
\emph{Ours} & {67.4} & \textbf{41.4} \\
\emph{Ours*}& \textbf{68.9} & {41.3} \\
\midrule
\textbf{Cityscapes} \\
\emph{Supervised} & - & \underline{77.3}\\
\emph{K-means} & 75.5 & 34.2 \\
\emph{Ours} & {86.1} & {46.7} \\
\emph{Ours*} & \textbf{88.3} & \textbf{52.8} \\
  \bottomrule
   \end{tabular}
\vspace{1ex}
\caption{Comparisons on semantic segmentation performance using the widely adopted evaluation protocol~\cite{strudel2021segmenter}. In particular, the background class in the Pascal Context is included for evaluation and the results on Cityscapes cover $19$ categories.  The pre-trained models with ViT-S architecture are used. Supervised results for the two datasets rely on DeepLabv3+~\cite{chen2018encoder} using ResNet-101~\cite{he2016deep} and Xception-65~\cite{chollet2017xception} backbones, respectively.}
\label{table:3}
\vspace{-0.2cm}
\end{table}
 
After that, we assess our method on ADE20K, one of the most challenging semantic segmentation datasets that contain $150$ fine-grained semantic categories. 
To tackle a large number of ground-truth semantic classes, we set $K$ to $512$ with $\beta$ as $0.04$. The training lasts for $20$ epochs, given that the training set is relatively large. 
We use the evaluation protocol of \cite{strudel2021segmenter}. 
We include the K-means baseline based on our own implementation for a fair comparison.

\begin{figure}[t]
\begin{center}
\includegraphics[width=\linewidth]{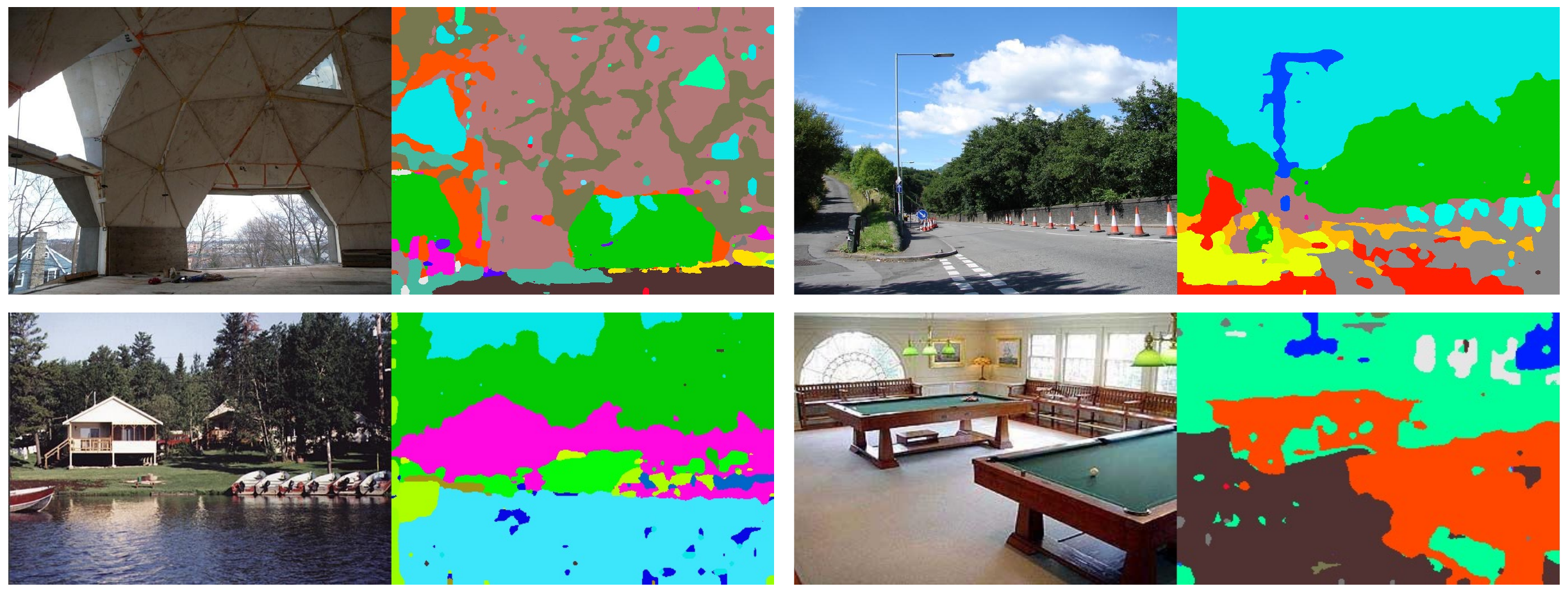}
\end{center}
   \caption{Visualization of the unsupervised semantic segmentation results of our method on ADE20K. In each pair, the left refers to the input image, and the right refers to the segmentation result. As shown, our method can yield reasonable pixel groups for images containing complex structures. }
\label{fig:4}
\end{figure}

The results are displayed in \cref{table:4}. As shown, our methods are still superior to the K-means baseline, especially regarding pixel accuracy. 
The best mIoU is $23.6\%$, which is much lower than the corresponding results on Pascal Context and Cityscapes. 
This probably stems from the large number of semantic categories in ADE20K---the used pre-trained models are not trained with fine-grained objectives (e.g., losses defined on patches), so they cannot ensure the semantic richness of the extracted dense features, which makes our method tend to generate clusters that conjoin multiple fine-grained semantic categories (see the lamp and the streetlight in \cref{fig:4}). 
One potential solution to this problem is to fine-tune the pre-trained models with patch-wise self-supervised learning loss and then invoke the proposed spectral clustering pipeline. 


\begin{table}[t]
  \centering
  \begin{tabular}{lcc@{}}
  \toprule
Method  & \multicolumn{1}{c}{\quad\quad\textbf{Acc.} (\%)\quad\quad}
& \multicolumn{1}{c}{\textbf{mIoU} (\%)}\\ 
\midrule
\emph{K-means} & 50.7 & 19.2 \\
\emph{Ours} & \textbf{63.3} & {21.6} \\
\emph{Ours*} & {62.5} & \textbf{23.6} \\
  \bottomrule
   \end{tabular}
\vspace{1ex}
\caption{Comparisons on unsupervised semantic segmentation performance on ADE20K using the evaluation protocol of \cite{strudel2021segmenter}. The pre-trained model with ViT-S architecture is used.}
  \label{table:4}
\end{table}

\subsection{Ablation Studies}
\label{sec:abl}
In this subsection, we present a comprehensive study of several key hyper-parameters in our method. 
We evaluate using the same setting as in \cref{sec:sota}. 
We consider the ViT-S architecture for per-trained models given its superior performance testified in previous results. 

\begin{table}[t]
  \centering
  \setlength{\tabcolsep}{8pt}
    \begin{tabular}{lcccc}
  \toprule
$K$  & \multicolumn{1}{c}{64} & \multicolumn{1}{c}{128} & \multicolumn{1}{c}{256} & \multicolumn{1}{c}{512}\\ 
\midrule
{Acc.} (\%) & 67.1 & 71.6 & 70.4 & 71.1 \\
{mIoU} (\%) & 27.8 & 33.2 & \underline{38.8} & 37.9 \\
  \bottomrule
   \end{tabular}
\vspace{1ex}
\caption{Performance comparison of different $K$ based on a pre-trained ViT-S model on Pascal Context validation set.} 
  \label{table:5}
\end{table}

\begin{table}[t]
  \centering
  \begin{tabular}{lccccc}
  \toprule
$\alpha$  & \multicolumn{1}{c}{0} & \multicolumn{1}{c}{0.1} & \multicolumn{1}{c}{0.3} & \multicolumn{1}{c}{0.5}  & \multicolumn{1}{c}{0.7}\\ 
\midrule
Pascal Context & 38.6 & 37.9 & 38.8 & \underline{38.9} & 38.0\\
Cityscapes & 27.2 & 26.7 & \underline{28.2} & 28.1 & 27.9 \\
  \bottomrule
   \end{tabular}
\vspace{1ex}
\caption{Mean IoU (\%) comparison of different trade-off coefficients $\alpha$ based on a pre-trained ViT-S model.} 
  \label{table:6}
\end{table}

\noindent \textbf{The output dimension $K$.} $K$ determines the number of eigenfunctions to learn and hence the dimension of the spectral embedding. 
It also implicitly connects to the number of semantic classes uncovered automatically during training. 
In previous experiments, we keep $K$ larger than the number of ground-truth semantic classes.
Here we perform an ablation study on $K$ on Pascal Context to reflect the necessity of doing so. 
The results are summarized in \cref{table:5}, which indicates that a value of $K$ of at least $256$ is necessary to achieve superior performance in terms of mIoU scores.
A larger $K$ yields more expressive representations that capture subtleties yet at the cost of increased computational overhead. 
Moreover, it should be noted that the NeuralEF technique may fail to uncover the eigenfunctions with small eigenvalues~\cite{deng2022neuralef}. 
Therefore, we suggest selecting a moderate value of $K$ in practice.

\noindent \textbf{The trade-off parameter $\alpha$.} The proposed spectral clustering workflow is compatible with any kernel function that captures plausible relationships between image patches. 
Currently, the used kernel is a weighted sum of the normalized adjacency defined on features from pre-trained models and that defined on down-sampled pixels. 
To verify the robustness of our method to the trade-off parameter $\alpha$, we perform an ablation study on it and report the results in \cref{table:6}. 
We include results for both Pascal Context and Cityscapes for a thorough investigation.
As shown, the mIoU on the validation data does not vary significantly w.r.t. $\alpha$, and limiting $\alpha$ to $[0.3, 0.5]$ can lead to superior results. 
Note that when $\alpha = 0$, i.e., we only use the features from pre-trained models to construct the graph kernel, the mIoU drops only slightly, indicating that the features from pre-trained models can retain most information on the neighborhood relationship of raw pixels. 
This also presents an opportunity for enhancing performance by improving the graph defined on down-sampled image pixels.

\noindent \textbf{The trade-off parameter $\beta$.} 
The trade-off parameter $\beta$ in ~\cref{eq:1} for learning neural eigenfunctions affects the empirical convergence. 
To investigate whether our method is sensitive to the choice of $\beta$, in \cref{table:7} we ablate the influence of $\beta$ in  both ``Ours'' and ``Ours*'' approaches on the Pascal Context data.
We observe that the validation mIoU remains rather stable across $\beta$ ranging from $0.04$ to $0.16$, which confirms the robustness of our method to $\beta$. 

\begin{table}[t]
  \centering
    \begin{tabular}{lccc}
  \toprule
$\beta$  & \multicolumn{1}{c}{0.04} & \multicolumn{1}{c}{0.08} & \multicolumn{1}{c}{0.16}\\ 
\midrule
\emph{Ours} & 38.8 & 38.8 & \underline{39.0} \\
\emph{Ours*} & 39.0 & \underline{39.6} & 39.3 \\
  \bottomrule
   \end{tabular}
\vspace{1ex}
\caption{Mean IoU (\%) comparison of different $\beta$ based on a pre-trained ViT-S model on Pascal Context validation set.} 
  \label{table:7}
\end{table}

\noindent \textbf{Zero-shot transfer.} 
Due to its unsupervised clustering nature, our method does not necessarily rely on target images for training. 
In this spirit, we perform an initial study where the training is conducted on ImageNet but the evaluation is conducted on both Pascal Context and Cityscapes. 
This forms a zero-shot transfer paradigm for unsupervised semantic segmentation. 
The training lasts for $5$ epochs under the same setting in previous studies. 
We report the results in \cref{table:8}. 
As a reference, the corresponding mIoU of MaskCLIP and ReCo on Cityscapes are $10.0$ and $19.3$ respectively~\cite{shin2022reco}, thus our method provides competitive performance. 
Nonetheless, the mIoUs for all the methods are much worse than those in previous studies. 
This is probably because images from ImageNet mostly contain clear foregrounds and single objects, which is in sharp contrast to the complex scene images in Pascal Context and Cityscapes, thus the learned neural eigenfunctions struggle to generalize. 
A potential remedy to this problem is to train on more realistic datasets to reduce the transfer gap.

\begin{table}[t]
  \centering
    \begin{tabular}{lcc@{}}
  \toprule
  & \multicolumn{1}{c}{\quad\quad{Acc.} (\%)\quad\quad}
& \multicolumn{1}{c}{{mIoU} (\%)}\\ 
\midrule
{Pascal Context}  & {55.8} & {15.2} \\
{Cityscapes} & {81.2} & {18.5} \\
  \bottomrule
   \end{tabular}
\vspace{1ex}
\caption{Zero-shot transfer results of our method. The training is performed on ImageNet based on the pre-trained ViT-S model. } 
  \label{table:8}
\end{table}
\section{Conclusion}
This work establishes an end-to-end NN-based pipeline for spectral clustering for unsupervised semantic segmentation. 
To achieve that, we build a connectivity graph over image patches using information from both pre-trained models and raw pixels and employ neural eigenfunctions to produce spectral embeddings corresponding to suitable graph kernels. 
We further quantize the output of the neural eigenfunctions to obtain clustering assignments without resorting to an explicit grouping step. 
After training, our method can generalize to novel test data easily and reliably. 
The reliance on pre-trained models gives our method good training efficiency and sufficient expressiveness. 
Extensive results confirm its superior performance over competing baselines. 

One limitation is that, like most clustering-based methods, our method needs to be exposed to ground-truth semantic masks to match clustering assignments to semantic segments. 
Introducing text prompts to guide clustering is a potential solution and deserves future investigation. 

{\small
\bibliographystyle{ieee_fullname}
\bibliography{egbib}

\begin{thebibliography}{10}\itemsep=-1pt

\bibitem{abdal2021labels4free}
Rameen Abdal, Peihao Zhu, Niloy~J Mitra, and Peter Wonka.
\newblock Labels4free: Unsupervised segmentation using stylegan.
\newblock In {\em Proceedings of the IEEE/CVF International Conference on
  Computer Vision}, pages 13970--13979, 2021.

\bibitem{arandjelovic2019object}
Relja Arandjelovi{\'c} and Andrew Zisserman.
\newblock Object discovery with a copy-pasting gan.
\newblock {\em arXiv preprint arXiv:1905.11369}, 2019.

\bibitem{bengio2003learning}
Yoshua Bengio, Pascal Vincent, Jean-Fran{\c{c}}ois Paiement, O Delalleau, M
  Ouimet, and N LeRoux.
\newblock Learning eigenfunctions of similarity: linking spectral clustering
  and kernel pca.
\newblock Technical report, Technical Report 1232, Departement d’Informatique
  et Recherche Oprationnelle~…, 2003.

\bibitem{chan2021comprehensive}
Lyndon Chan, Mahdi~S Hosseini, and Konstantinos~N Plataniotis.
\newblock A comprehensive analysis of weakly-supervised semantic segmentation
  in different image domains.
\newblock {\em International Journal of Computer Vision}, 129:361--384, 2021.

\bibitem{chen2018encoder}
Liang-Chieh Chen, Yukun Zhu, George Papandreou, Florian Schroff, and Hartwig
  Adam.
\newblock Encoder-decoder with atrous separable convolution for semantic image
  segmentation.
\newblock In {\em Proceedings of the European conference on computer vision
  (ECCV)}, pages 801--818, 2018.

\bibitem{chen2019unsupervised}
Micka{\"e}l Chen, Thierry Arti{\`e}res, and Ludovic Denoyer.
\newblock Unsupervised object segmentation by redrawing.
\newblock {\em Advances in neural information processing systems}, 32, 2019.

\bibitem{cho2021picie}
Jang~Hyun Cho, Utkarsh Mall, Kavita Bala, and Bharath Hariharan.
\newblock Picie: Unsupervised semantic segmentation using invariance and
  equivariance in clustering.
\newblock In {\em Proceedings of the IEEE/CVF Conference on Computer Vision and
  Pattern Recognition}, pages 16794--16804, 2021.

\bibitem{chollet2017xception}
Fran{\c{c}}ois Chollet.
\newblock Xception: Deep learning with depthwise separable convolutions.
\newblock In {\em Proceedings of the IEEE conference on computer vision and
  pattern recognition}, pages 1251--1258, 2017.

\bibitem{cordts2016cityscapes}
Marius Cordts, Mohamed Omran, Sebastian Ramos, Timo Rehfeld, Markus Enzweiler,
  Rodrigo Benenson, Uwe Franke, Stefan Roth, and Bernt Schiele.
\newblock The cityscapes dataset for semantic urban scene understanding.
\newblock In {\em Proceedings of the IEEE conference on computer vision and
  pattern recognition}, pages 3213--3223, 2016.

\bibitem{deng2009imagenet}
Jia Deng, Wei Dong, Richard Socher, Li-Jia Li, Kai Li, and Li Fei-Fei.
\newblock Imagenet: A large-scale hierarchical image database.
\newblock In {\em 2009 IEEE conference on computer vision and pattern
  recognition}, pages 248--255. Ieee, 2009.

\bibitem{deng2022neural}
Zhijie Deng, Jiaxin Shi, Hao Zhang, Peng Cui, Cewu Lu, and Jun Zhu.
\newblock Neural eigenfunctions are structured representation learners.
\newblock {\em arXiv preprint arXiv:2210.12637}, 2022.

\bibitem{deng2022neuralef}
Zhijie Deng, Jiaxin Shi, and Jun Zhu.
\newblock Neuralef: Deconstructing kernels by deep neural networks.
\newblock {\em arXiv preprint arXiv:2205.00165}, 2022.

\bibitem{donath1973lower}
William~E Donath and Alan~J Hoffman.
\newblock Lower bounds for the partitioning of graphs.
\newblock {\em IBM Journal of Research and Development}, 17(5):420--425, 1973.

\bibitem{dosovitskiy2020image}
Alexey Dosovitskiy, Lucas Beyer, Alexander Kolesnikov, Dirk Weissenborn,
  Xiaohua Zhai, Thomas Unterthiner, Mostafa Dehghani, Matthias Minderer, Georg
  Heigold, Sylvain Gelly, et~al.
\newblock An image is worth 16x16 words: Transformers for image recognition at
  scale.
\newblock {\em arXiv preprint arXiv:2010.11929}, 2020.

\bibitem{fiedler1973algebraic}
Miroslav Fiedler.
\newblock Algebraic connectivity of graphs.
\newblock {\em Czechoslovak mathematical journal}, 23(2):298--305, 1973.

\bibitem{hamilton2022unsupervised}
Mark Hamilton, Zhoutong Zhang, Bharath Hariharan, Noah Snavely, and William~T
  Freeman.
\newblock Unsupervised semantic segmentation by distilling feature
  correspondences.
\newblock {\em arXiv preprint arXiv:2203.08414}, 2022.

\bibitem{he2016deep}
Kaiming He, Xiangyu Zhang, Shaoqing Ren, and Jian Sun.
\newblock Deep residual learning for image recognition.
\newblock In {\em Proceedings of the IEEE conference on computer vision and
  pattern recognition}, pages 770--778, 2016.

\bibitem{huang2018weakly}
Zilong Huang, Xinggang Wang, Jiasi Wang, Wenyu Liu, and Jingdong Wang.
\newblock Weakly-supervised semantic segmentation network with deep seeded
  region growing.
\newblock In {\em Proceedings of the IEEE conference on computer vision and
  pattern recognition}, pages 7014--7023, 2018.

\bibitem{hwang2019segsort}
Jyh-Jing Hwang, Stella~X Yu, Jianbo Shi, Maxwell~D Collins, Tien-Ju Yang, Xiao
  Zhang, and Liang-Chieh Chen.
\newblock Segsort: Segmentation by discriminative sorting of segments.
\newblock In {\em Proceedings of the IEEE/CVF International Conference on
  Computer Vision}, pages 7334--7344, 2019.

\bibitem{jang2016categorical}
Eric Jang, Shixiang Gu, and Ben Poole.
\newblock Categorical reparameterization with gumbel-softmax.
\newblock {\em arXiv preprint arXiv:1611.01144}, 2016.

\bibitem{ji2019invariant}
Xu Ji, Joao~F Henriques, and Andrea Vedaldi.
\newblock Invariant information clustering for unsupervised image
  classification and segmentation.
\newblock In {\em Proceedings of the IEEE/CVF International Conference on
  Computer Vision}, pages 9865--9874, 2019.

\bibitem{katharopoulos-et-al-2020}
A. Katharopoulos, A. Vyas, N. Pappas, and F. Fleuret.
\newblock Transformers are rnns: Fast autoregressive transformers with linear
  attention.
\newblock In {\em Proceedings of the International Conference on Machine
  Learning (ICML)}, 2020.

\bibitem{ke2021universal}
Tsung-Wei Ke, Jyh-Jing Hwang, and Stella~X Yu.
\newblock Universal weakly supervised segmentation by pixel-to-segment
  contrastive learning.
\newblock {\em arXiv preprint arXiv:2105.00957}, 2021.

\bibitem{khoreva2017simple}
Anna Khoreva, Rodrigo Benenson, Jan Hosang, Matthias Hein, and Bernt Schiele.
\newblock Simple does it: Weakly supervised instance and semantic segmentation.
\newblock In {\em Proceedings of the IEEE conference on computer vision and
  pattern recognition}, pages 876--885, 2017.

\bibitem{kingma2014adam}
Diederik~P Kingma and Jimmy Ba.
\newblock Adam: A method for stochastic optimization.
\newblock {\em arXiv preprint arXiv:1412.6980}, 2014.

\bibitem{krahenbuhl2011efficient}
Philipp Kr{\"a}henb{\"u}hl and Vladlen Koltun.
\newblock Efficient inference in fully connected crfs with gaussian edge
  potentials.
\newblock {\em Advances in neural information processing systems}, 24, 2011.

\bibitem{kuhn1955hungarian}
Harold~W Kuhn.
\newblock The hungarian method for the assignment problem.
\newblock {\em Naval research logistics quarterly}, 2(1-2):83--97, 1955.

\bibitem{law2017deep}
Marc~T Law, Raquel Urtasun, and Richard~S Zemel.
\newblock Deep spectral clustering learning.
\newblock In {\em International conference on machine learning}, pages
  1985--1994. PMLR, 2017.

\bibitem{lloyd1982least}
Stuart Lloyd.
\newblock Least squares quantization in pcm.
\newblock {\em IEEE transactions on information theory}, 28(2):129--137, 1982.

\bibitem{melas2022deep}
Luke Melas-Kyriazi, Christian Rupprecht, Iro Laina, and Andrea Vedaldi.
\newblock Deep spectral methods: A surprisingly strong baseline for
  unsupervised semantic segmentation and localization.
\newblock In {\em Proceedings of the IEEE/CVF Conference on Computer Vision and
  Pattern Recognition}, pages 8364--8375, 2022.

\bibitem{mottaghi2014role}
Roozbeh Mottaghi, Xianjie Chen, Xiaobai Liu, Nam-Gyu Cho, Seong-Whan Lee, Sanja
  Fidler, Raquel Urtasun, and Alan Yuille.
\newblock The role of context for object detection and semantic segmentation in
  the wild.
\newblock In {\em Proceedings of the IEEE conference on computer vision and
  pattern recognition}, pages 891--898, 2014.

\bibitem{ng2001spectral}
Andrew Ng, Michael Jordan, and Yair Weiss.
\newblock On spectral clustering: Analysis and an algorithm.
\newblock {\em Advances in neural information processing systems}, 14, 2001.

\bibitem{ouali2020autoregressive}
Yassine Ouali, C{\'e}line Hudelot, and Myriam Tami.
\newblock Autoregressive unsupervised image segmentation.
\newblock In {\em Computer Vision--ECCV 2020: 16th European Conference,
  Glasgow, UK, August 23--28, 2020, Proceedings, Part VII 16}, pages 142--158.
  Springer, 2020.

\bibitem{pfau2018spectral}
David Pfau, Stig Petersen, Ashish Agarwal, David~GT Barrett, and Kimberly~L
  Stachenfeld.
\newblock Spectral inference networks: Unifying deep and spectral learning.
\newblock {\em arXiv preprint arXiv:1806.02215}, 2018.

\bibitem{radford2021learning}
Alec Radford, Jong~Wook Kim, Chris Hallacy, Aditya Ramesh, Gabriel Goh,
  Sandhini Agarwal, Girish Sastry, Amanda Askell, Pamela Mishkin, Jack Clark,
  et~al.
\newblock Learning transferable visual models from natural language
  supervision.
\newblock In {\em International Conference on Machine Learning}, pages
  8748--8763. PMLR, 2021.

\bibitem{ridnik2021imagenet21k}
Tal Ridnik, Emanuel Ben-Baruch, Asaf Noy, and Lihi Zelnik-Manor.
\newblock Imagenet-21k pretraining for the masses, 2021.

\bibitem{scholkopf2005kernel}
Bernhard Sch{\"o}lkopf, Alexander Smola, and Klaus-Robert M{\"u}ller.
\newblock Kernel principal component analysis.
\newblock In {\em Artificial Neural Networks—ICANN'97: 7th International
  Conference Lausanne, Switzerland, October 8--10, 1997 Proceeedings}, pages
  583--588. Springer, 2005.

\bibitem{shi2000normalized}
Jianbo Shi and Jitendra Malik.
\newblock Normalized cuts and image segmentation.
\newblock {\em IEEE Transactions on Pattern Analysis and Machine Intelligence},
  22(8):888--905, 2000.

\bibitem{shin2022reco}
Gyungin Shin, Weidi Xie, and Samuel Albanie.
\newblock Reco: Retrieve and co-segment for zero-shot transfer.
\newblock {\em arXiv preprint arXiv:2206.07045}, 2022.

\bibitem{strudel2021segmenter}
Robin Strudel, Ricardo Garcia, Ivan Laptev, and Cordelia Schmid.
\newblock Segmenter: Transformer for semantic segmentation.
\newblock In {\em Proceedings of the IEEE/CVF International Conference on
  Computer Vision}, pages 7262--7272, 2021.

\bibitem{van2021unsupervised}
Wouter Van~Gansbeke, Simon Vandenhende, Stamatios Georgoulis, and Luc Van~Gool.
\newblock Unsupervised semantic segmentation by contrasting object mask
  proposals.
\newblock In {\em Proceedings of the IEEE/CVF International Conference on
  Computer Vision}, pages 10052--10062, 2021.

\bibitem{voynov2021object}
Andrey Voynov, Stanislav Morozov, and Artem Babenko.
\newblock Object segmentation without labels with large-scale generative
  models.
\newblock In {\em International Conference on Machine Learning}, pages
  10596--10606. PMLR, 2021.

\bibitem{weiss2008spectral}
Yair Weiss, Antonio Torralba, and Rob Fergus.
\newblock Spectral hashing.
\newblock {\em Advances in neural information processing systems}, 21, 2008.

\bibitem{rw2019timm}
Ross Wightman.
\newblock Pytorch image models.
\newblock \url{https://github.com/rwightman/pytorch-image-models}, 2019.

\bibitem{yang2019deep}
Xu Yang, Cheng Deng, Feng Zheng, Junchi Yan, and Wei Liu.
\newblock Deep spectral clustering using dual autoencoder network.
\newblock In {\em Proceedings of the IEEE/CVF conference on computer vision and
  pattern recognition}, pages 4066--4075, 2019.

\bibitem{zhang2020self}
Xiao Zhang and Michael Maire.
\newblock Self-supervised visual representation learning from hierarchical
  grouping.
\newblock {\em Advances in Neural Information Processing Systems},
  33:16579--16590, 2020.

\bibitem{zhou2019semantic}
Bolei Zhou, Hang Zhao, Xavier Puig, Tete Xiao, Sanja Fidler, Adela Barriuso,
  and Antonio Torralba.
\newblock Semantic understanding of scenes through the ade20k dataset.
\newblock {\em International Journal of Computer Vision}, 127:302--321, 2019.

\bibitem{zhou2022extract}
Chong Zhou, Chen~Change Loy, and Bo Dai.
\newblock Extract free dense labels from clip.
\newblock In {\em European Conference on Computer Vision (ECCV)}, 2022.

\end{thebibliography}
}

\newpage 
\appendix

\section{Extra Visualization Results}

\begin{figure}[h]
\begin{center}
\includegraphics[width=\linewidth]{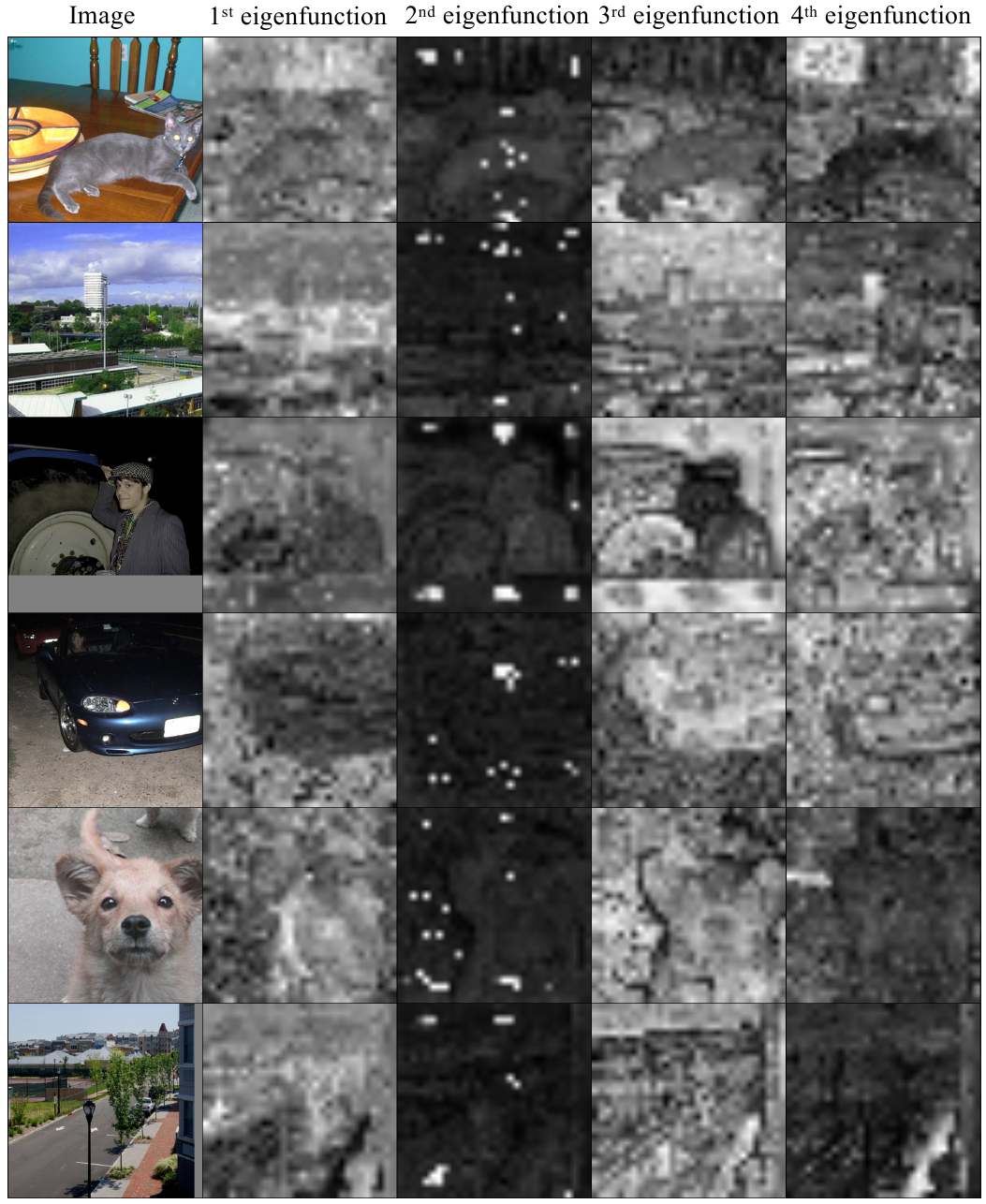}
\end{center}
\vspace{-.5ex}
   \caption{Visualization of the learned neural eigenfunctions on Pascal Context.}
  \vspace{-.5ex}
\label{fig:12}
\end{figure}

\cref{fig:12} shows the learned neural eigenfunctions on Pascal Context. 
We clarify that the plots correspond to the outputs of the neural eigenfunctions under standard $[0, 1]$ normalization (bilinear up-sampling is also used). 
Softmax transformation is not applied. 
We see that different eigenfunctions respond to different input patterns actively. 
Combining them is obviously beneficial to solving edge detection and image segmentation problems. 

\begin{figure}[t!]
\begin{center}
\includegraphics[width=\linewidth]{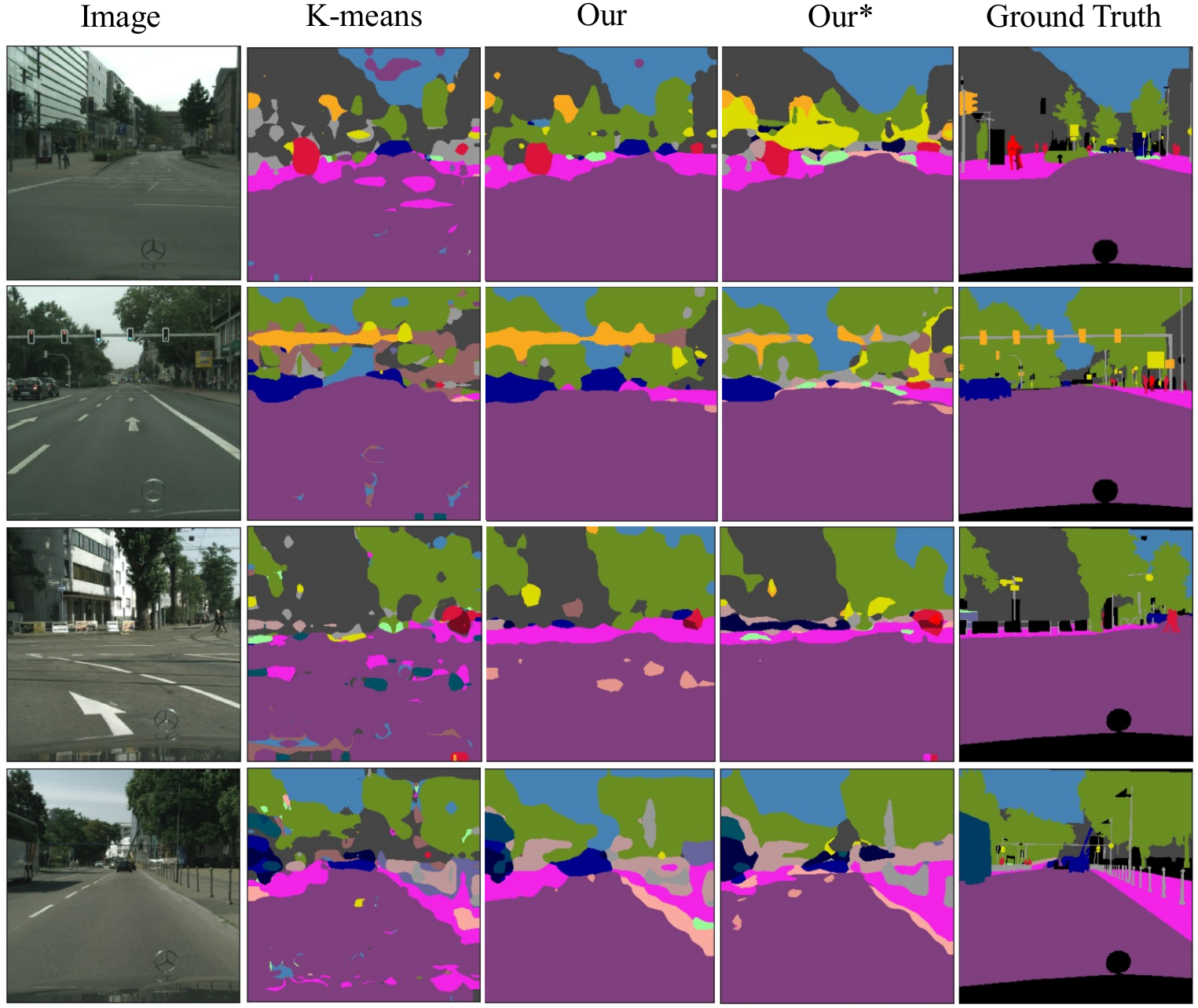}
\end{center}
\vspace{-.5ex}
   \caption{Visualization of the unsupervised semantic segmentation results on Cityscapes.}
  \vspace{-.5ex}
\label{fig:10}
\end{figure}

\cref{fig:10} shows some qualitative results of the proposed methods on Cityscapes.
Notably, our method can detect multiple semantic categories in the same image and the generated object boundaries are sharp. 

\end{document}